\documentclass[10pt,twocolumn,letterpaper,pagenumbers]{article}

\usepackage[pagenumbers]{wacv} % To force page numbers, \eg for an arXiv version

\usepackage{graphicx}
\usepackage{amsmath}
\usepackage{amssymb}
\usepackage{booktabs}
\usepackage{fancyhdr}

\usepackage[pagebackref,breaklinks,colorlinks]{hyperref}

\usepackage[capitalize]{cleveref}
\crefname{section}{Sec.}{Secs.}
\Crefname{section}{Section}{Sections}
\Crefname{table}{Table}{Tables}
\crefname{table}{Tab.}{Tabs.}

\def\wacvPaperID{668} % *** Enter the WACV Paper ID here
\def\confName{WACV}
\def\confYear{2025}

\begin{document}
\newcommand\blfootnote[1]{%
  \begingroup
  \renewcommand\thefootnote{}\footnote{#1}%
  \addtocounter{footnote}{-1}%
  \endgroup
}
\fancypagestyle{firstpage}
{
    \fancyhead[L]{\textbf{Accepted to appear in the proceedings of WACV 2025}}    
    \fancyhead[R]{}
}
\newcommand{\magma}[0]{\texttt{MAGMA}}
\newcommand{\lreg}[0]{$\mathcal{L}_{\text{Reg}}$}

% \title{\LaTeX\ Author Guidelines for \confName~Proceedings}
\title{\texttt{MAGMA:} Manifold Regularization for MAEs}

\author{Alin Dondera$^{*,1}$, Anuj Singh$^{*,1,2}$, Hadi Jamali-Rad$^{1,2}$\\
$^1$Delft University of Technology (TU Delft), The Netherlands \\
$^2$Shell Global Solutions International B.V., Amsterdam, The Netherlands \\
{\tt\small a.e.dondera@student.tudelft.nl, \{a.r.singh, h.jamalirad\}@tudelft.nl}
% \and
}

% \author{Alin Dondera$^{*}$\\
% Delft University of Technology\\
% {\tt\small a.e.dondera@student.tudelft.nl} \\
% {\tt\small dondera.alin@gmail.com}
% \and
% Anuj Singh$^{*}$\\
% Delft University of Technology\\
% Shell Global Solutions International B.V.\\
% {\tt\small a.r.singh@tudelft.nl} \\
% {\tt\small anuj.singh2@shell.com}
% \and
% Hadi Jamali-Rad\\
% Delft University of Technology\\
% Shell Global Solutions International B.V.\\
% {\tt\small h.jamalirad@tudelft.nl} \\
% {\tt\small hadi.jamali-rad@shell.com}
% }
\maketitle

%%%%%%%%% ABSTRACT
\begin{abstract}
  Masked Autoencoders (MAEs) are an important divide in self-supervised learning (SSL) due to their independence from augmentation techniques for generating positive (and/or negative) pairs as in contrastive frameworks. Their masking and reconstruction strategy also nicely aligns with SSL approaches in natural language processing. Most MAEs are built upon Transformer-based architectures where visual features are not regularized as opposed to their convolutional neural network (CNN) based counterparts, which can potentially hinder their performance. To address this, we introduce \magma{}, a novel batch-wide layer-wise regularization loss applied to representations of different Transformer layers. We demonstrate that by plugging in the proposed regularization loss, one can significantly improve the performance of MAE-based models. We further demonstrate the impact of the proposed loss on optimizing other generic SSL approaches (such as VICReg and SimCLR), broadening the impact of the proposed approach. Our code base can be found here: \url{https://github.com/adondera/magma}   
\end{abstract}
% \pagestyle{fancy}
% \fancyhead{Accepted to appear in the proceedings of WACV2025}
\thispagestyle{firstpage}
\blfootnote{* equal contribution}

%%%%%%%%% BODY TEXT
\section{Introduction}
\label{sec:intro}

\begin{figure*}
  \centering
  \includegraphics[height=4.9cm]{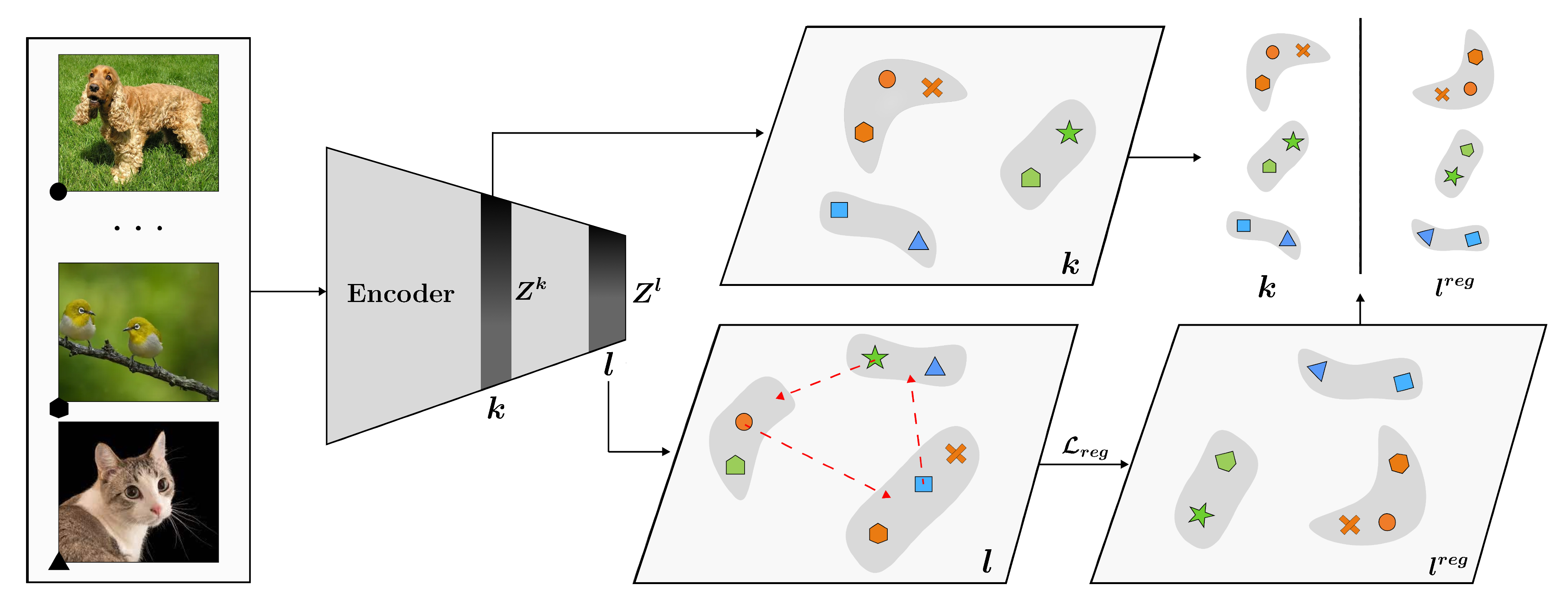}
  \caption{Visualization for the proposed regularization loss \magma{} with MAE: \magma{} penalizes representations that are close in the latent space of intermediate layer $k$ but far apart in layer $l$ latent space. This induces a regularization effect across different layers that preserves inter-sample and intra-batch relationships thus enforcing consistency in the latent representation space. Note that we demonstrate this for MAE based pre-training with a transformer encoder-decoder architecture such as ViT.}
  \label{fig:method-visualization}
\end{figure*}

Self-supervised learning has made significant progress over the recent years by producing results on par with supervised baselines \cite{bardes2021vicreg, grill2020bootstrap, zbontar2021barlow, chen2020simple, caron2020unsupervised, caron2021emerging}, thus rendering it as a promising paradigm for learning representations without access to labels. Many notable approaches in self-supervised learning such as contrastive learning \cite{chen2020simple}, clustering-based methods \cite{caron2020unsupervised}, redundancy minimization \cite{bardes2021vicreg, zbontar2021barlow} and distillation-based methods \cite{grill2020bootstrap} aim to learn representations that generalize well by avoiding degenerate solutions and representational collapse by utilising a joint embedding architecture to enforce consistency between representations of different image-views. Inspired by natural language processing (NLP), Masked Autoencoders (MAE) approach the task of self-supervised pre-training by a conceptually simple idea of masking a portion of the input data to then learn to predict the removed content. Specifically, this is applied to images by masking a very large portion (eg. $75$\%) of their content by replacing it with random patches. This creates a challenging pretext task for image representation learning that requires the neural network to develop a holistic understanding beyond low-level image statistics \cite{he2022masked}. By masking a large part of the image and processing only the unmasked region, MAEs provide a computationally efficient way of pre-training large-scale vision transformers such as ViT-B/H/S \cite{he2022masked, dosovitskiy2020image}. However, due to the lack of an objective that optimizes for contrasting negative pairs of images, the features learnt by MAE pre-training require large amounts of labeled data to be fine-tuned for satisfactory downstream task performance \cite{lehner2023contrastive}. Moreover, deep architectures such as convolutional neural networks are designed with inherent regularization characteristics such as translation invariance, equivariance, and parameter sharing that are relevant to learning information-rich features from images for multiple vision-oriented tasks. On the other hand, ViT-based architectures operate on patches of images and lack these aforementioned regularization characteristics in their feature extraction process. In an ideal scenario, a well-trained network should exhibit a crucial property: if two similar inputs are fed into the network, their resulting outputs should also be close together. This principle ensures that the network learns robust representations that capture the underlying structure of the data, not just random noise or specific details. Deviations from this principle can indicate the network is overly sensitive to small input variations, leading to poor generalization performance on unseen data. One way to enforce this behavior is through manifold regularization, which aims to guide the model toward learning smoother representations aligned with the intrinsic data geometry \cite{belkin2006manifold}. To this end, we introduce \magma{}, a novel batch-wide loss that regularizes representations across multiple different layers of a feature extractor. Our extensive experiments, ablations, and analyses empirically demonstrate improved downstream image classification performance on MAE-based baselines by simply plugging in the proposed regularization loss during the pre-training phase. To corroborate the general applicability and broader impact of \magma{}, we demonstrate improved and on-par performance of other generic SSL approaches such as VICReg \cite{bardes2021vicreg} and SimCLR \cite{chen2020simple} when pre-trained with our proposed loss.

\section{Related Work}

\textbf{Self-supervised learning.}
Self-supervised learning (SSL) is crucial for overcoming the limitations of traditional supervised learning, which requires vast amounts of expensive, hand-labeled data. By automatically generating labels from the data itself, self-supervised techniques enable models to learn meaningful representations from unlabeled data, reducing our reliance on manual annotation. A common strategy in self-supervised learning (SSL) is to exploit augmentation invariance. By enforcing similarity between augmented views of the same image, these methods aim to learn representations that are robust to common image transformations. To prevent model collapse, various techniques are employed, such as negative sampling \cite{chen2020simple}, cluster assignments, \cite{caron2020unsupervised}, feature decorrelation \cite{bardes2021vicreg, zbontar2021barlow}, or asymmetric architectures \cite{grill2020bootstrap}.

\textbf{Masked Autoencoders.}
The success of self-supervised learning in Natural Language Processing (NLP), particularly with masked language modeling techniques in models like BERT \cite{devlin2018bert} has inspired analogous developments within computer vision. Masked Autoencoders (MAEs) \cite{he2022masked} take the idea of masking and apply it to an autoencoder structure with a pixel-level reconstruction loss. This results in impressive performance across various downstream tasks \cite{pang2022masked, zhang2022mask, chen2023masked, zhou2022self}. Other similar works include BEiT \cite{bao2021beit}, SimMIM \cite{xie2022simmim}, and iBOT \cite{zhou2021ibot}, with close connections to contrastive learning \cite{kong2019mutual, zhang2022mask}.

\textbf{Manifold regularization.} At the core of \magma{} lies the seminal piece of work of \cite{belkin2006manifold}. The authors provide a geometrically intuitive and novel semi-supervised learning framework that leverages the underlying geometry of data distributions under the assumption that two points close together on the manifold (i.e., similar in the true underlying structure of the data), should have their corresponding target outputs also be similar. This idea has been successfully applied in deep learning across of variety of tasks, such as speech recognition \cite{tomar2014manifold, tomar2016graph}, NLP \cite{yonghe2019refining, li2021topic} and vision \cite{jie2015manifold, jin2020manifold, shaham2018spectralnet, hu2018robust}, showcasing its usefulness in the general setting.
\magma{} extends the concept of manifold regularization to the self-supervised setting, guiding internal network transformations to promote smoother, more generalizable representations. While \cite{shaham2018spectralnet} explores a similar direction, their approach relies on Siamese networks to explicitly calculate similarity measures between input images. In contrast, our regularization operates directly on the representations generated within the network, offering a more tightly integrated self-supervised mechanism.
\section{Method: \texttt{MAGMA}}
\label{sec:method}

% Tricky writing this mathematical formulation nicely. Maybe check some of the more theoretical works from LeCun for notation, etc.
Given an unlabeled dataset $\mathcal{D}_u$ with samples $x \in \mathcal{D}_u$ our goal is to train an encoder $f_{\theta}$ with $L$ layers to produce information-rich representations in a self-supervised fashion. During inference, the parameters of the encoder are frozen $\theta$ and a linear layer is trained in a supervised fashion. This procedure is known as linear probing and is the commonly adopted setup in self-supervised learning (SSL) literature. We denote a batch of $B$ samples as $\mathcal{B}$. In this setting, our goal is to apply a batch-level regularization loss in a layer-wise fashion on a set of layers $\mathcal{K} \subseteq [L]$:
\begin{equation}
\label{eq:genericloss}
\mathcal{L}(\mathcal{B}, \mathcal{K}; \theta) = \mathcal{L}_{SSL}(\mathcal{B}; \theta) + \lambda \mathcal{L}_{Reg}(\mathcal{B}, \mathcal{K}; \theta), \nonumber
\end{equation}
where the first term denotes a standard self-supervised learning loss, and $\lambda$ is a weighting parameter between the two terms. While any set of layers can in practice be adopted for such a regularization, we demonstrate later on that applying this on an intermediate and the last layer $\mathcal{K} = \{l, L\}$ would yield the maximum impact. Notably, this is applied only at the pretraining phase. 

\subsection{Batch-Wide Layer-Wise Manifold Reg.}
We denote the representation output of layer $l \in [L]$ of $f_\theta$ for input image $i \in \mathcal{B}$ as $Z_i^{(l)}$. Inspired by \cite{belkin2006manifold}, we propose to apply the following batch-wide layer-wise regularization term to enforce consistency among the output representations of the selected layers:
\begin{small}
\begin{equation} \label{eq:2}
    \mathcal{L}_{\text{Reg}}(\mathcal{B}, \mathcal{K}; \theta) = \frac{1}{B^2} \sum_{k, l \in \mathcal{K}} \, \sum_{i, j \in [B]} w(Z_{i}^{(k)}, Z_{j}^{(k)}) \cdot || Z_{i}^{(l)} - Z_{j}^{(l)} ||^2 
\end{equation}
\end{small}

\noindent where $w(.) : \mathbb{R}^D \times \mathbb{R}^D \rightarrow \mathbb{R}$ can be any similarity kernel, $D$ being the size of the vectorized version of $Z$. In our study, we employ the Radial Basis Function (RBF) kernel due to its favorable properties as discussed in \cite{belkin2006manifold}. Thus, we have:
\begin{equation} \label{eq:4}
    w(Z_{i}^{(k)}, Z_{j}^{(k)}) = \exp{(\frac{-||Z_{i}^{(k)} - Z_{j}^{(k)}||^2}{2\sigma})} \nonumber
\end{equation}
\vspace{-0.2cm}

\noindent where $\sigma$ is a free parameter. We choose $\sigma^2 = \texttt{var}(d_{ij})$, with $d_{ij} = ||Z_{i}^{(k)} - Z_{j}^{(k)}||^2$, following the approach in \cite{rodriguez2020embedding} for enhanced training stability. Dynamically adjustment of $\sigma$ this way ensures our regularization adapts to the spread of features inside a batch: the more spread out the features are (i.e. higher $\sigma$) the wider the influence of the RBF kernel. Conversely, a lower spread would result in a more focused kernel (focusing on finer, more local distinctions). Note that in Eq.~\ref{eq:2}, layer $k$ is considered as the reference layer and layer $l$ is regularized accordingly. More concretely, if two instances ($Z_{i}$ and $Z_{j}$) have closer representations in the manifold space of layer $k$ (leading to higher $w(Z_{i}^{(k)}, Z_{j}^{(k)})$), but are far apart in the manifold space of layer $l$, $\mathcal{L}_{\text{Reg}}$ would heavily penalize them, as a result pulling them closer in the regularized manifold. We illustrate later on that in practice these regularizations would not only regularize layer $l$ but also all the previous layers.

The regularization loss in Eq.~\ref{eq:2} can be reformulated in terms of the Laplacian matrix $L$ determined by all pairs of instances $(Z_{i}^{(k)}, Z_{j}^{(k)})$ in a batch, and is defined as follows:
% \begin{equation} \label{eq:4}
%     \mathcal{L}_{\text{Reg}}(\mathcal{B}, \mathcal{K}; \theta) = \frac{1}{B^2} \sum_{i, j = 1}^{B} W_{ij} \cdot || Z_{i}^{(k)} - Z_{j}^{(k)} ||^2.
% \end{equation}
%
\begin{equation} \label{eq:6}
    \mathcal{L}_{\text{Reg}}(\mathcal{B}, \mathcal{K}; \theta) = \frac{1}{B^2} \texttt{Trace}(Z^{(l)T} L Z^{(l)}), \nonumber
\end{equation}
We make use of the normalized Laplacian for better stability during training, defined as follows:
\begin{equation} \label{eq:7}
    L = D^{-\frac{1}{2}}WD^{-\frac{1}{2}},\: D_{ii}=\sum_jW_{ij} = \sum_jw(Z_{i}^{(k)}, Z_{j}^{(k)}), \nonumber
\end{equation}

\textbf{Application to Transformers.} For the sake of generality, we have so far formulated the problem so that it would be readily applied to any layered neural network architecture. Even though, we have only observed significant impact on ViT based architectures. The only difference for ViT based architectures is that per layer $l$ we would have $P$ patches each of which returning a representation $Z^{(l)}_{i,p}$, $\forall p \in [P]$, where the image level representation would simply be the average of all those representations $Z^{(l)}_{i} = \sum_p Z^{(l)}_{i,p}$. The reason behind this averaging strategy is that applying the regularization over the representations of individual patches across different images is not ideal due to patch noise and lack of global context. This may result in irrelevant computations since similar patches within an image already share context through self-attention. 

\section{Impact of Architectures and Pretraining} \label{sec:pretrain}
The proposed regularization can be seamlessly incorporated into various self-supervised methods, with the caveat that the chosen architecture and pretraining approach play an important role in determining the efficacy of the regularization. The inherent characteristics of CNN-based architectures can diminish the impact of regularization. For instance parameter sharing, translation invariance and equivariance in CNNs, which facilitates the reuse of learned features across various input regions, can result in reduced regularization impact. In contrast, Transformers lack these specific characteristics, potentially making them more suitable for this regularization.

The nature of the pretraining method significantly influences the impact of regularization. Contrastive methods (\eg, SimCLR, MoCo), clustering approaches (SwAV), distillation techniques (DINO, BYOL), and InfoMax/Dimension Contrastive methods all aim to bring representations of augmented views of the same image closer together, essentially performing a task related to our proposed regularization. Therefore, the proposed regularization will have a diminished impact on these methods. On the other hand, Masked Autoencoders (MAE)'s exhibits a generative nature, by randomly masking large portions of an image and reconstructing the missing pixels. Since this process is applied individually, it is also not sharing any information between representations within a batch. These characteristics make it better suited for the regularization term. As a result, our study will primarily focus on MAEs as they align well with the objectives of our proposed regularization approach.

\section{Experiments}

\begingroup
\begin{table*}
\setlength{\tabcolsep}{6pt} % Default value: 6pt
  \caption{Linear probing accuracy and k-nn accuracy (k=10) of models pretrained and evaluated on the given datasets. Adding our proposed regularization term to the baseline method generally increases performance.
  }
  \vspace{-0.3cm}
  \label{tab:linear-probing-main}
  \centering
  \begin{tabular}{@{}l|cccccccc@{}}
    \hline
    & \multicolumn{2}{c}{CIFAR-100} & \multicolumn{2}{c}{STL-10} & \multicolumn{2}{c}{Tiny-ImageNet} & \multicolumn{2}{c}{ImageNet-100} \\
    Method & linear & knn & linear & knn & linear & knn & linear & knn \\
    \hline
    MAE  & 38.2 & 36.6 & 66.5 & 62.0 & 17.8 & 17.7 & 58.0 & 47.5\\
    M-MAE (ours)  & \textbf{43.3} & \textbf{40.7} & \textbf{71.0} & \textbf{65.9} & \textbf{20.9} & \textbf{20.5} & \textbf{69.0} & \textbf{49.8} \\
    \hline
    U-MAE & 45.3 & 45.9 & 74.9 & 72.1 & 21.5 & 19.0 & 69.5 & 56.8 \\
    MU-MAE (ours) & \textbf{46.4} & \textbf{46.4} & \textbf{75.6} & \textbf{73.0} & \textbf{25.2} & \textbf{23.9} & \textbf{73.4} & \textbf{60.1} \\
    \hline
    SimCLR  & 62.8 & 58.7 & \textbf{90.4} & \textbf{86.9} & \textbf{50.9} & 43.5 & 67.8 & 65.3 \\
    M-SimCLR (ours) & \textbf{63.2} & \textbf{59.4} & \textbf{90.5} & \textbf{86.9} & \textbf{51.0} & \textbf{44.6} & \textbf{68.7} & \textbf{65.6} \\
    \hline
    VICReg  & 63.6 & 60.8 & \textbf{87.4} & \textbf{84.5} & 45.2 & \textbf{40.5} & 68.4 & 62.1 \\
    M-VICReg (ours) & \textbf{64.7} & \textbf{61.9} & \textbf{87.4} & \textbf{84.5} & \textbf{45.8} & \textbf{40.5} & \textbf{70.4} & \textbf{65.1} \\
  \hline
  \end{tabular}
\end{table*}
\endgroup

\begingroup
\begin{table*}
\setlength{\tabcolsep}{6pt} % Default value: 6pt
  \caption{Linear probing (LP) and fine-tuning (FT) accuracy of models pretrained on ImageNet-100.}
  \vspace{-0.3cm}
  \label{tab:ft}
  \centering
  {\begin{tabular}{@{}l|cccccccc@{}}
    \hline
    & \multicolumn{2}{c}{CIFAR-100} & \multicolumn{2}{c}{STL-10} & \multicolumn{2}{c}{Tiny-ImageNet} & \multicolumn{2}{c}{ImageNet-100} \\
    Method & LP & FT & LP & FT & LP & FT & LP & FT \\
    \hline
    MAE  & 31.5 & \textbf{76.9} & 67.8 & 82.2 & 27.8 & \textbf{63.1} & 58.0 & 79.8\\
    M-MAE (ours)  & \textbf{51.6} & 75.6 & \textbf{84.8} & \textbf{85.5} & \textbf{43.1} & \textbf{63.1} & \textbf{69.0} & \textbf{80.6} \\
    \hline
  \end{tabular}}
\end{table*}
\endgroup

\begin{table}[h]
  \caption{Linear probing results on different versions of ImageNet for MAE \ U-MAE with and without regularization.} 
  % All models were trained by equally sampling X\% of each class from ImageNet for training.}
  \vspace{-0.3cm}
  \label{tab:smaller-regime}
  \centering
  \begin{tabular}{@{}ccccc@{}}
    \toprule
    Method & 1\% & 5\% & 10\% & 100\% \\
    \midrule
    MAE  & 0.5 & 2.7 & 4.3 & 50.5\\
    M-MAE (ours) & \textbf{1.6} & \textbf{8.8} & \textbf{13.2} & \textbf{51.0} \\
    \midrule
    U-MAE & 2.9 & 15.4 & 25.6 & 55.3 \\
    MU-MAE (ours) & \textbf{4.9} & \textbf{20.1} & \textbf{29.6} & \textbf{57.4} \\
  \bottomrule
  \end{tabular}
\end{table}

Our goal in this section is to evaluate the impact of adding our regularization term on top of pre-existing SSL methods, both quantitatively and qualitatively. We aim to address the following questions: \\ 
\textbf{[Q1]} How does $\mathcal{L}_\textup{Reg}$ influence downstream classification?\\
\textbf{[Q2]} What is the effect of $\mathcal{L}_\textup{Reg}$ on the training dynamics? \\
\textbf{[Q3]} What are the important hyperparameters of the proposed regularization?  \\ 
\textbf{[Q4]} Is the impact of $\mathcal{L}_\textup{Reg}$ on representations qualitatively noticeable?

\textbf{Benchmark Datasets.} We evaluate our proposed regularization on commonly adopted datasets for the downstream task of image classification, namely, CIFAR100 \cite{krizhevsky2009learning}, STL-10 \cite{coates2011analysis}, Tiny-Imagenet \cite{le2015tiny}, and ImageNet-100 \cite{tian2020contrastive}. This selection of datasets provides various challenges in terms of data resolution, number of classes, and overall complexity of context presented in the sample image. By testing across diverse datasets, we showcase the robustness and generalizability of our proposed regularization.

\textbf{Baseline methods.} We evaluate the efficacy of our proposed regularization on several SSL methods (to demonstrate its versatility), with an emphasis on MAE for the reasons discussed in \cref{sec:pretrain}. This includes U-MAE \cite{zhang2022mask}, an improvement over the baseline MAE addressing dimensional collapse with an additional regularization term. Additionally, we investigate the impact on two other widely adopted SSL baselines: SimCLR \cite{chen2020simple} and VICReg \cite{bardes2021vicreg}.

% We've also considered approaches like BYOL \cite{grill2020bootstrap} or DINO \cite{caron2021emerging}, but the results were not successful. We hypothesize that is because the training dynamics of both are sensitive to the choice of hyperparameters. 

\textbf{Implementation details}. We focus on the impact of regularization by keeping the architectural and hyperparameter choice intact throughout the experimentation, except for the ablation studies. For low(er)-resolution datasets (CIFAR100, STL-10, and Tiny-ImageNet) we use a ViT-Tiny backbone, while for ImageNet-100, we use ViT-Base. We select the best-performing hyperparameter setting for each baseline method and add our regularization on top of it. For our regularization, we tune three parameters: the regularization weight $\lambda$, the number of warmup epochs $e_{\textup{st}}$, and the duration of the regularizer $e_{\textup{dur}}$.  More details on the choice of (hyper)parameters can be found in the supplementary material. Notably, for all the other baselines, we present the reproduction results in one optimized pipeline, which in almost all cases leads to performances over the originally reported results in \cite{bardes2021vicreg, chen2020simple, he2022masked, zhang2022mask}. 

{\bf Evaluation protocol.} For our main results, we follow the commonly adopted protocol in SSL, based on freezing the network encoder after the pretraining phase and training a linear layer on top of it in a supervised fashion. For all baselines, we train for $100$ epochs using SGD, using a learning rate of $0.1$ with decay at steps $60$ and $80$, and a batch size of $256$. In addition, we also evaluate the k-nearest neighbours ($k$NN) classification accuracy using $k=10$ and a Euclidean distance measure.
% \vspace{0.5cm}

\begin{figure*}[t]
  \centering
  \begin{subfigure}{0.47\linewidth}
    \includegraphics[width=1\linewidth]{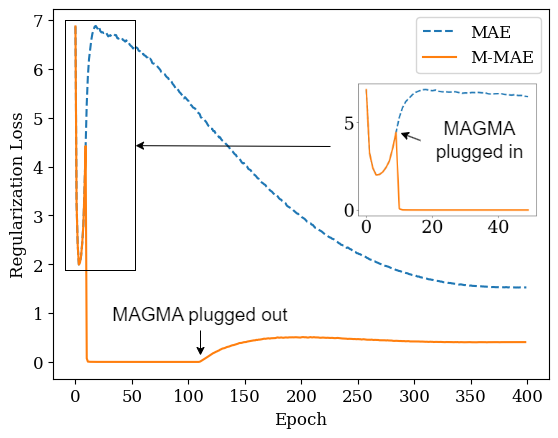}
      \caption{Regularization loss tracked throughout pretraining for MAE and M-MAE, on ImageNet-100}
      \label{fig:reg-plot-baseline}
  \end{subfigure}
 \hfill
  \begin{subfigure}{0.47\linewidth}
      \includegraphics[width=1\linewidth]{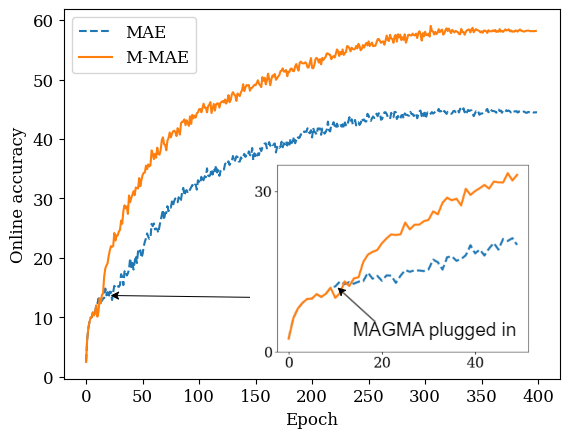}
      \caption{Online accuracy tracked throughout the pretraining phase for MAE and M-MAE, on ImageNet-100.}
      \label{fig:online-accuracy}
  \end{subfigure}
  \caption{(a) The regularization loss showed for MAE and M-MAE. For MAE we calculate the loss without backpropagating. For M-MAE, we apply the loss after 10 warmup epochs, and take it out after 100 epochs. (b) The online accuracy was obtained by training a linear layer on the representations produced by the encoder throughout pretraining. The accuracy slightly drops for M-MAE when the regularization kicks in but increases at a significantly higher rate compared to MAE.}
  \label{fig:reg-impact}
\end{figure*}

\subsection{[AQ1] Comparison Against Other Baselines} 
\Cref{tab:linear-probing-main} summarizes the results of applying \texttt{MAGMA} on top of the aforementioned four baselines (MAE, U-MAE, SimCLR, and VICReg). We pretrain and evaluate on the same datasets to showcase the robustness of \magma{} over various pretraining scenarios. As can be seen, our proposed approach offers significant improvements across all four datasets by outperforming baseline MAE, both in the linear setting ($+5.1$\% on CIFAR-100, $+4.5$\% on STL10, $+2.9$\% on Tiny-ImageNet and $+11$\% on ImageNet-100), as well for $k$NN one (\textbf{+4.1}\% on CIFAR-100, \textbf{+3.9}\% on STL-10, \textbf{+2.8}\% on Tiny-ImageNet, and \textbf{+2.3}\% on ImageNet 100). For U-MAE, while the improvements are still significant, except for Tiny-ImageNet, they are smaller in magnitude. Going beyond MAEs, the results on SimCLR and VICReg, show some marginal improvement opening the door for further investigation and broader impact. To further demonstrate the impact of \magma{}'s enhanced self-supervised representation learning at the pretraining phase, we evaluate classification performance of MAE and M-MAE on different datasets by linear-probing and fine-tuning in \Cref{tab:ft}. As can be seen, M-MAE significantly outperforms MAE across all \vspace{0.4mm} datasets when using linear probing for downstream dataset classification, offering improvements up to $20\%$. Full fine-tuning (especially in low-data regimes) can lead to overfitting to the target dataset \cite{zhang2022mask}, completely defeating the purpose and ruling out the impact of our regularization. This is what we also observe in \Cref{tab:ft}, where M-MAE offers similar results as compared to the baseline MAE. To assess the effectiveness of MAGMA in lower-data regimes and the commonly adopted large-scale SSL pretraining, we conducted linear probing evaluations on various versions of ImageNet, where we equally sampled $1\%$, $5\%$, $10\%$ and the complete $100\%$ split from each class of the original dataset for training. \vspace{0.5mm} Table \ref{tab:smaller-regime} demonstrates that both M-MAE and MU-MAE significantly outperform their respective baselines (MAE and U-MAE) across all partial sampling ratios by atleast $1.5$\% and up to $8.9$\%, as well as for the full ImageNet-1k dataset by $0.5$\% and $2.1$\%. This shows the efficacy of \magma{} in scenarios not only with limited data but also for large-scale pre-training. 

\subsection{[AQ2] Training Dynamics}\label{sec:training-dynamics}

\Cref{fig:reg-plot-baseline} illustrates the value of the proposed loss term ($\mathcal{L}_{Reg}$) throughout training epochs. The dashed line illustrates the scenario in which $\mathcal{L}_{Reg}$ is evaluated but not backpropagated. This curve manifest signs of instability (lack of consistency) in the manifold space of representations (for selected layers $11$ and $12$). The solid curve shows the impact of backpropagating $\mathcal{L}_{Reg}$ (applying \magma{} at epoch $10$) where a sudden change of behavior is apparent upon the introduction of $\mathcal{L}_{Reg}$ in the optimization. The fact that the $\mathcal{L}_{Reg}$ drops drastically instead of ascending (dashed line) after being introduced, together with the stability of the loss after removal (at epoch $110$), as well as the consistently better online accuracy of M-MAE as seen in \cref{fig:online-accuracy}, could potentially suggest that the optimization is now steered in a different direction, leading to an overall significantly better performance. Based on this, we hypothesize that the $e_{\textup{st}}$ parameter is best set around the point when the \lreg{} loss would start increasing.

% \textbf{Plot showing no effect on the reconstruction loss should also be present}.

% \textbf{Loss curves}.

% \textbf{U-MAE and MU-MAE}. Can show two plots here. How does the uniformity loss from MU-MAE influence the regularization loss, and vice-versa.

% \begin{figure}
%     \centering
%     \includegraphics[width=0.5\linewidth]{loss_curve_per_layer.png}
%     \caption{As you go deeper into the model, the bigger the loss becomes. Later representations tend to break our desired property}
%     \label{fig:enter-label}
% \end{figure}

\begin{figure}
    \centering
    \includegraphics[width=1.0\linewidth]{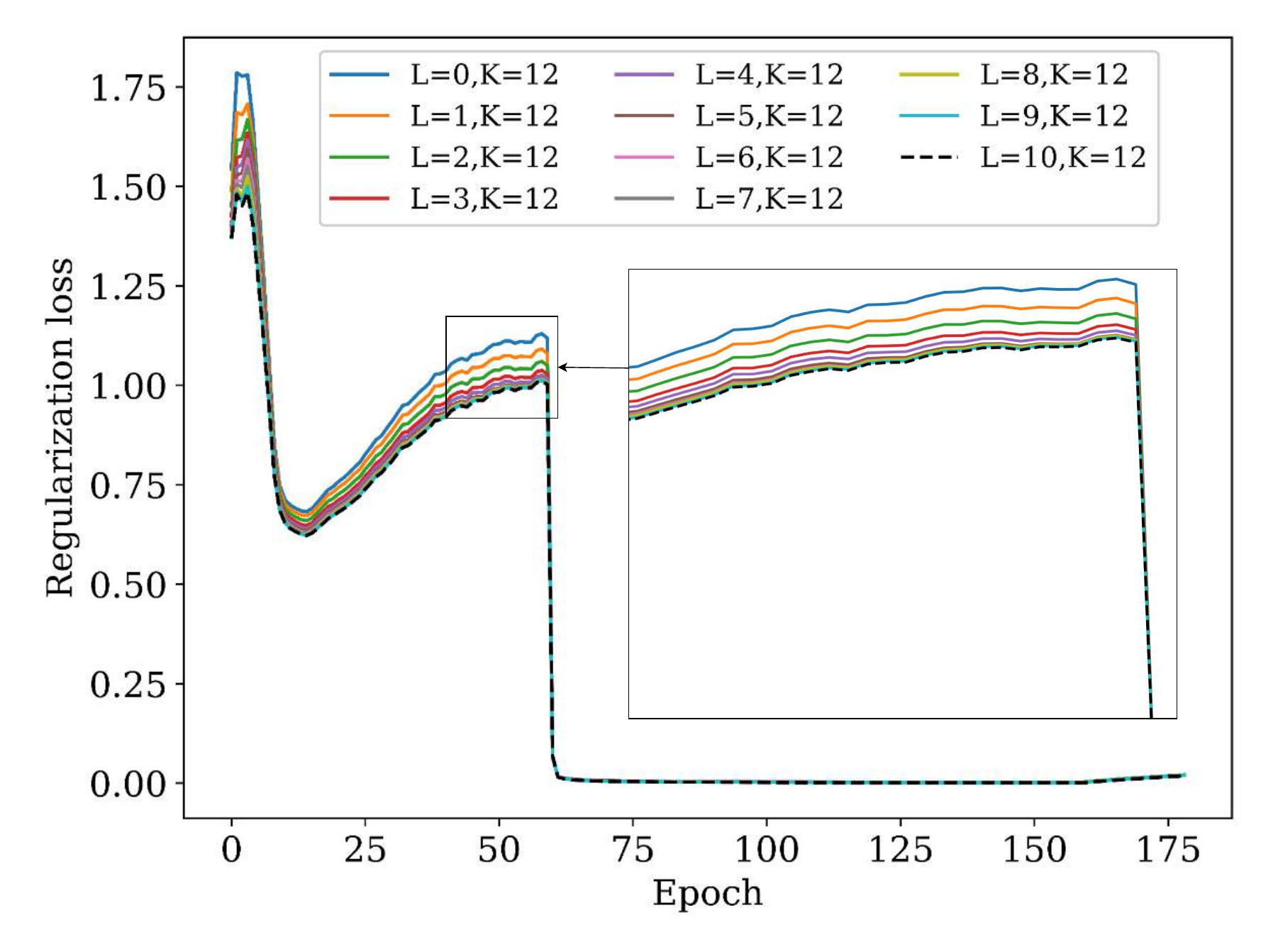}
    \vspace{-0.3cm}
    \caption{Effect of regularization. Implication: if the representations from any two layers are close, then the output representation will also be close.}
    \label{fig:diff-layers-reg}
    \vspace{-0.5cm}
\end{figure}

\textbf{Which layers to regularize on?}
We have run extensive experimentation to effectively select the target layers for applying \magma. It turns out regularizing the last layer with respect to the penultimate layer seems to have the maximum impact, in ViT based architecture. In \cref{fig:diff-layers-reg}, we demonstrate that choosing $k = 10$  ($11$-th later in ViT base architecture) as the reference and $l = 11$ (last year) not only leads to regularizing loss across the two layers, but also results in percolated impact through all the previous layers.

% Maybe appendix
% \textbf{Regularization scheme changes}. Applying the regularization at different layers and its impact.
% \begin{figure}
%     \centering
%     \includegraphics[width=0.5\linewidth]{loss_consecutive_layers.png}
%     \caption{Enter Caption}
%     \label{fig:enter-label}
% \end{figure}
% \textbf{Applying the method at the end}. Show whether performance can be improved by applying regularization for a couple of epochs at the end. Using ImageNet-100.

\begingroup
\setlength{\tabcolsep}{4pt}
\renewcommand{\arraystretch}{1.2} % Default value: 1
\begin{table*}[tb]
  \caption{Linear accuracy performance using different choices of hyperparameters for regularization. Results computed on ImageNet-100.}
  \label{tab:main-ablations}
  \centering
  \begin{tabular}{@{}l|cccc|cccc|cccc@{}}
  \hline
    $\lambda$ & \textit{1} & \textit{0.1} & \textit{0.01} & \textit{10} & 1 & 1 & 1 & 1 & 1 & 1 & 1 & 1 \\
    $e_{\textup{st}}$ & 10 & 10 & 10 & 10 & \textit{0} & \textit{2} & \textit{20} & \textit{50} & 10 & 10 & 10 & 10 \\
    $e_{\textup{dur}}$ & 100 & 100 & 100 & 100 & 100 & 100 & 100 & 100 & \textit{10} & \textit{50} & \textit{200} & \textit{390} \\
    % $\sigma$ & $\texttt{std}(d_{ij})$ & $\texttt{std}(d_{ij})$ & $\texttt{std}(d_{ij})$ & $\texttt{std}(d_{ij})$ & $\texttt{std}(d_{ij})$ & $\texttt{std}(d_{ij})$ & $\texttt{std}(d_{ij})$ & $\texttt{std}(d_{ij})$ & - \\
    \hline
    Accuracy & 69.0 & 60.0 & 54.6 & 67.9 & 68.2 & 69.0 & 65.5 & 62.7 & 68.5 & 68.8 & 69.0 & 68.5 \\
  \hline
  \end{tabular}
\end{table*}
\endgroup

\subsection{[AQ3] Ablations on Important Parameters}
We evaluate the impact of three pivotal regularization parameters (i) $\lambda$ in \cref{eq:genericloss}, (ii) the epoch at which \magma is applied, $e_{\textup{st}}$, and (iii) the duration over which the regularization is applied before being plugged out, $e_{\textup{dur}}$.  The results for the first three parameters are summarized in \cref{tab:main-ablations}. 

The regularization weight $\lambda$ directly controls the strength of the regularization effect in the overall optimization loss \cref{eq:genericloss}. Intuitively, lower weight for \lreg{} might not significantly impact the overall optimization, whereas higher weight could lead to an over-regularized optimization and a degraded performance. The results show a similar trend: lowering the weight to $0.1$ leads to a performance similar to the baseline ($+2$\%). Increasing the weight by a factor of $10$ reduces the gain slightly by $1\%$. Interestingly, reducing the weight to $0.01$ leads to lower downstream classification performance than the baseline. We hypothesize that this is because the regularization introduces a competing gradient signal which inadvertently hinders training performance.

The warm up period $e_{\textup{st}}$ allows the model to train for a few epochs without \lreg{} to help it establish a reasonable foundation for learning basic representations. This could prevent the regularization from overly restricting the model too early in the training process. As can be seen from \cref{tab:main-ablations}, a small number of epochs for $e_{\textup{st}}$ would already be enough for a maximal impact. Delaying this further seems to have an increasingly negative impact.  

Lastly, the duration parameter $e_{\textup{dur}}$, determines the amount of pressure put on the model to develop smooth and aligned representations across layers. We experiment with different values ranging from only $10$ epochs, up until the end of training (i.e. a duration of $390$ epochs). The results show that the impact of this parameter is less pronounced. There is a slight decrease in performance (by about $0.5\%$), for significantly lower or higher duration periods. It seems that applying \magma{} for a number of epochs already regularizes the representations across the network with a lingering impact from which point onward it can be plugged out without hampering the overall performance. As discussed earlier in \cref{sec:training-dynamics}, we hypothesize that this lingering impact is related to the adjusted optimization landscape as a result of applying the proposed regularization. 

To further investigate the sensitivity of \magma, we evaluate the performance by changing the backbone architecture starting from small to larger (ViT-S to ViT-L). As can be seen in \cref{tab:backbones}, increasing the capacity of the backbone results in considerable performance improvement in the baseline approaches (MAE and U-MAE) where the performance boosts decreases for changing the backbone from ViT-B to ViT-L. Interestingly, similarly significant boost can be observed on the \magma{} optimized baselines (M-MAE, MU-MAE), offering consistent improvement over the baselines.

% Regularization term effect cannot be too strong as it can then overly constrain the model. It can also not be too small as it will not give any significant gain. Apply it at the peak of the loss (verify this in ablation) for a fixed number of epochs (100). We notice that after stopping the regularization, the loss increases slightly as we plug it out but then remains constant. The more we use the regularization, the lesser the loss, but the more we constrain the model during training which can impact its generalization capabilities. Weight equal to 1 works ok. Too low and it doesn't give good performance. 

% \textbf{RBF Kernel Weight}. We use the inverse std. of features as the RBL kernel weight. The bigger their standard deviation, the more similar we will consider points. With a very high std. even points that are far apart will have some similarities. It would be interesting to see how this parameter influences the results. We can try to keep the standard deviation, but we can try and scale it by some number and ablate this.

\begin{table}[h]
  \caption{Linear probing with different ViTs on ImageNet-100.
  }
  \label{tab:backbones}
  \centering
  \begin{tabular}{@{}cccc@{}}
    \toprule
    Method & ViT-S & ViT-B & ViT-L\\
    \midrule
    MAE  & 46.8 & 57.9 & 60.6\\
    M-MAE (ours) & \textbf{61.2} & \textbf{69.2} & \textbf{73.9}\\
    \midrule
    U-MAE & 57.6 & 69.5 & \textbf{78.2} \\
    MU-MAE (ours) & \textbf{62} & \textbf{73.4} & \textbf{78.4} \\
  \bottomrule
  \end{tabular}
\end{table}

\begin{figure*}[t]
        \centering
        \begin{subfigure}[b]{0.45\textwidth}
            \centering
            \includegraphics[width=\textwidth]{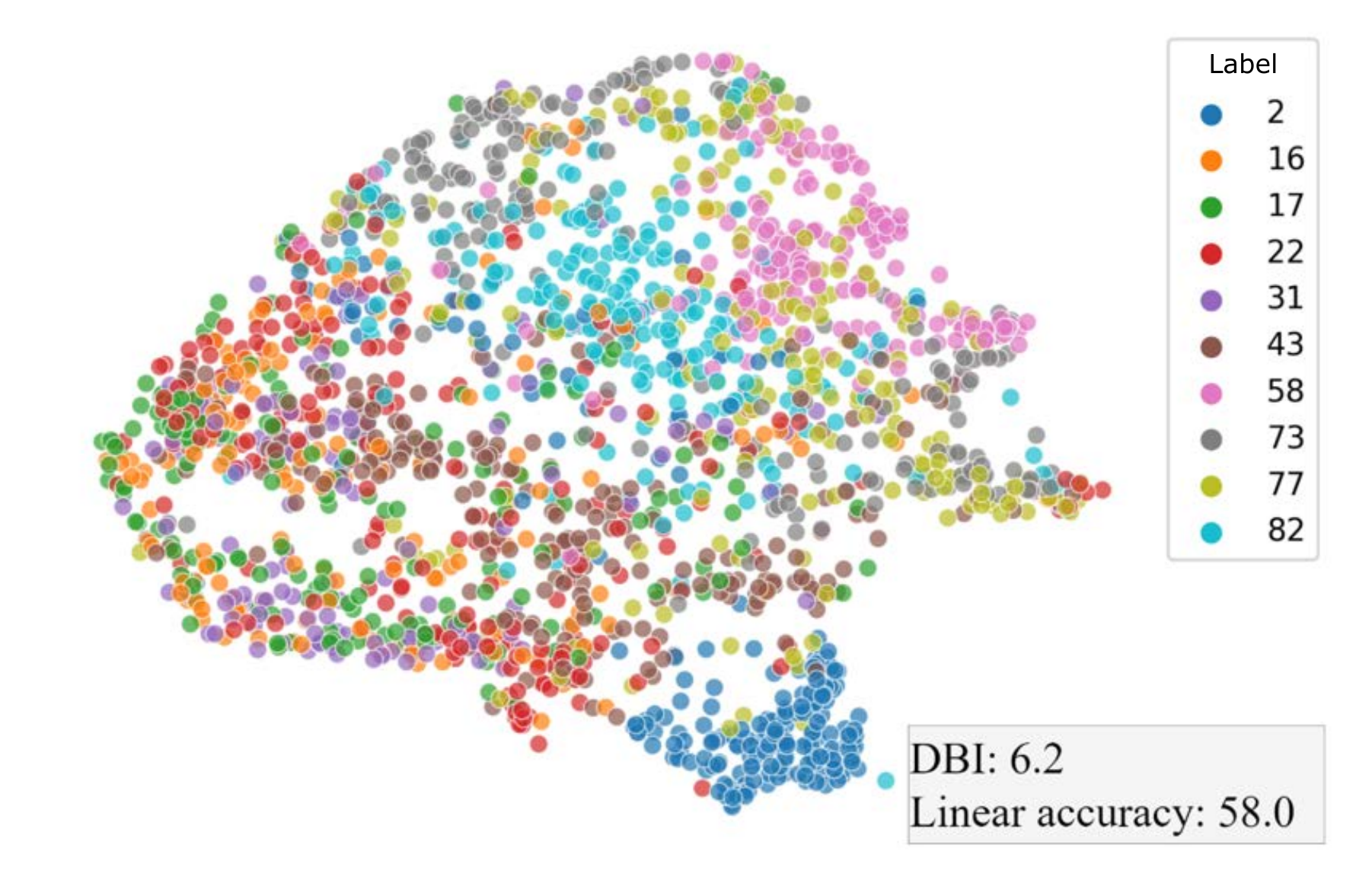}
            \caption[Network2]%
            {{\small MAE}}    
            \label{fig:pacmap-mae}
        \end{subfigure}
        \hfill
        \begin{subfigure}[b]{0.45\textwidth}  
            \centering 
            \includegraphics[width=\textwidth]{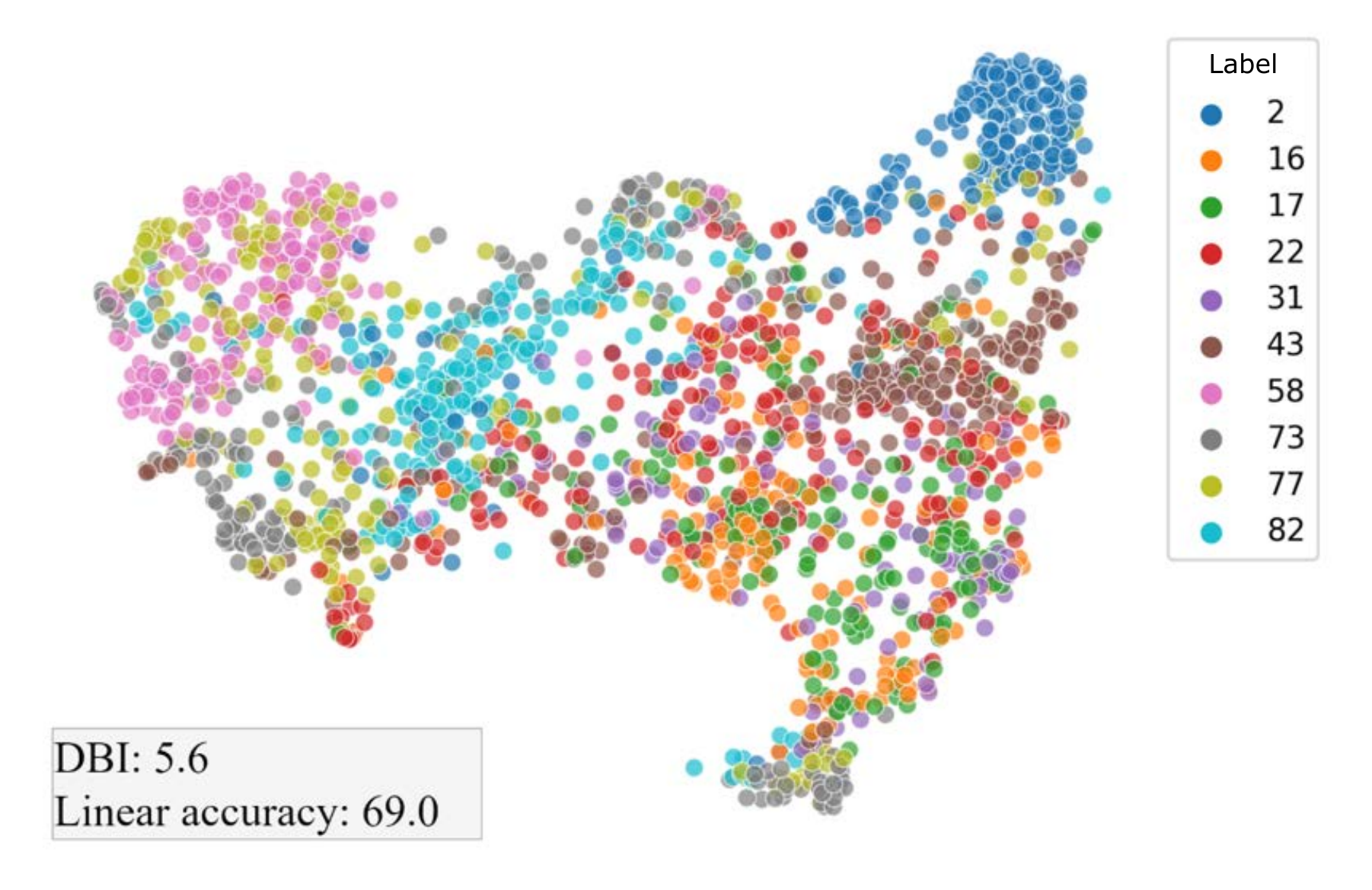}
            \caption[]%
            {{\small M-MAE}}    
            \label{fig:pacmap-m-mae}
        \end{subfigure}
        \vskip\baselineskip
        \begin{subfigure}[b]{0.45\textwidth}   
            \centering 
            \includegraphics[width=\textwidth]{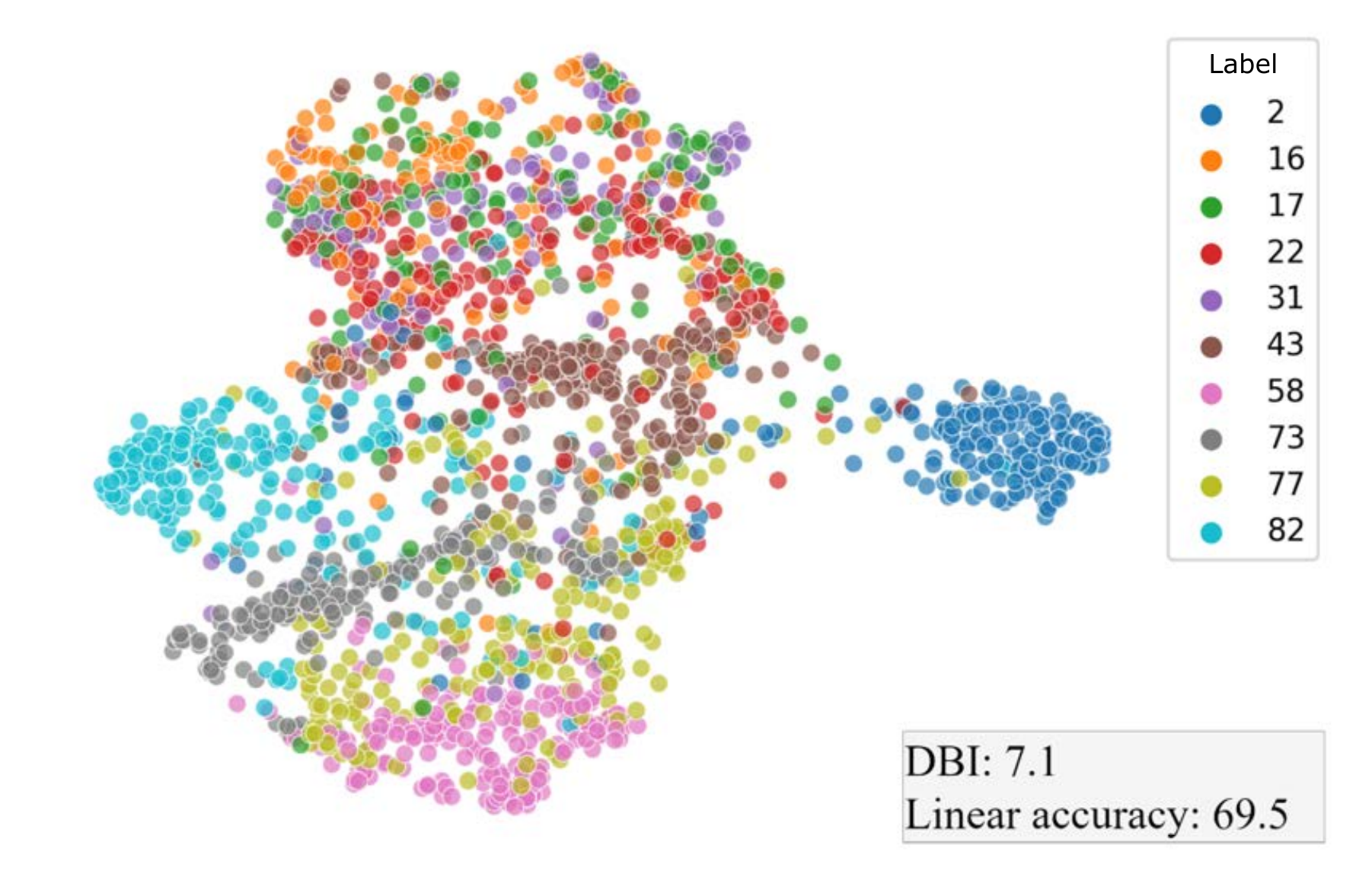}
            \caption[]%
            {{\small U-MAE}}    
            \label{fig:pacmap-u-mae}
        \end{subfigure}
        \hfill
        \begin{subfigure}[b]{0.45\textwidth}   
            \centering 
            \includegraphics[width=\textwidth]{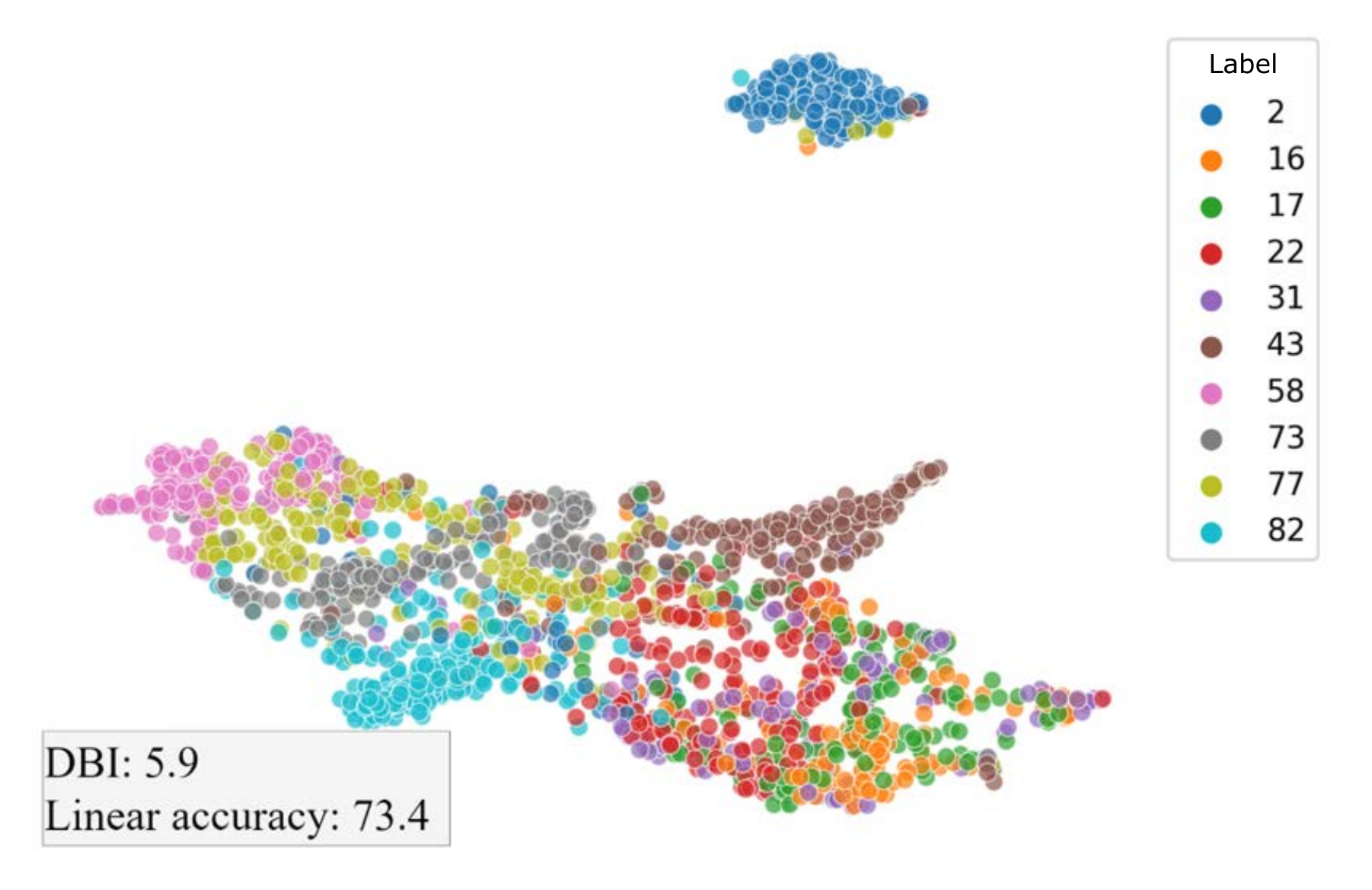}
            \caption[]%
            {{\small MU-MAE}}    
            \label{fig:pacmap-mu-mae}
        \end{subfigure}
        \caption
        {\small PaCMAP plots for MAE-based methods. Applying \magma{} on top of U-MAE leads to compact and well-defined clusters.} 
        \label{fig:pacmaps}
    \end{figure*}

\subsection{[AQ4] Qualitative Analysis}

\textbf{PacMAP.} To qualitatively assess the impact of our regularization, we visualize the representations of MAE, U-MAE, as well as their regularized version, M-MAE, and MU-MAE, on a random sample of 10 classes from ImageNet-100. We use PacMAP \cite{wang2021understanding} for dimensionality reduction. PaCMAP outperforms t-SNE \cite{van2008visualizing} and UMAP \cite{mcinnes2018umap} in preserving the global structure of high-dimensional data within visualizations. This means it more accurately reflects the large-scale relationships and patterns present in the original dataset. We include the linear accuracy, as well as the Davies-Bouldin Index (DBI) \cite{davies1979cluster} alongside the visualizations. DBI is a common metric used to evaluate clustering algorithms, where lower DBI scores indicate better separation between clusters and tighter groupings within clusters.

Results are shown in \cref{fig:pacmaps}. Comparing (a) MAE with (b) M-MAE, we observe a slight improvement in the clustering structure after applying \magma{}. The M-MAE representations exhibit tighter clusters with better class separation, leading to a higher linear accuracy (69.0 vs. 58.0) and a lower DBI (5.6 vs. 6.2). Similarly, comparing (c) U-MAE and (d) MU-MAE reveals that incorporating \magma{} into U-MAE  further refines the representation space. While U-MAE already improves upon the baseline, MU-MAE achieves even tighter clusters and greater inter-class separation, resulting in a further boost in accuracy (73.4) and a lower DBI (5.9). This highlights the complementary nature of MAGMA and U-MAE, where MAGMA enhances the already improved representations learned by U-MAE.

\textbf{PCA.} Inspired by \cite{amir2021deep}, we take our pretrained MAEs and extract features from each patch, and each layer (in this case, we isolate the \textit{key} features from the self-attention mechanism), apply PCA, and take the leading component. We use upsampling to obtain a heatmap of the same resolution as the original image. This provides a qualitative analysis of the quality of the intermediate representations learned by the models, showcasing the impact of the added regularization term. One visible pattern is the reduction in noise, specifically in the first and last layers, that M-MAE exhibits when compared to its MAE counterpart.

\begin{figure*}
  \vspace{-0.5cm}
  \centering
  \includegraphics[width=\textwidth]{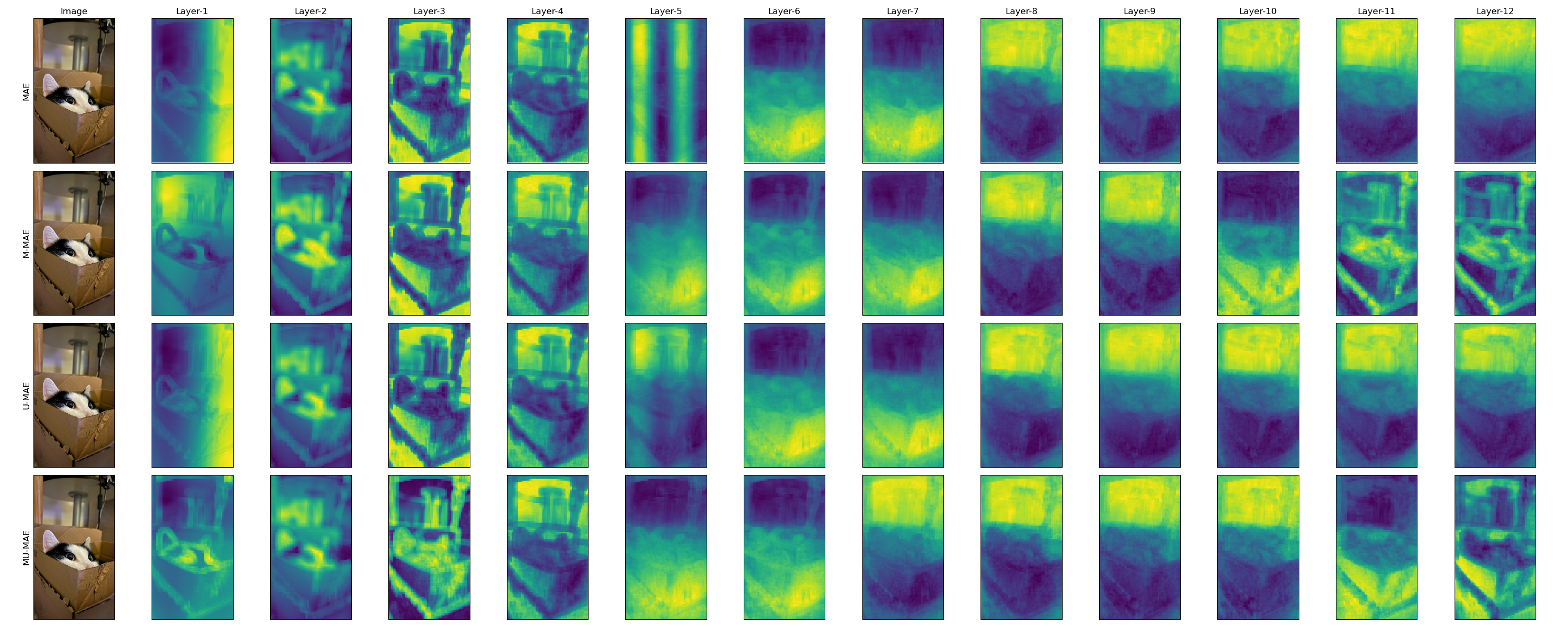}
    \vspace{-0.5cm}
  \caption{Visualization of PCA's leading component for features extracted from different layers of a ViT-B pretrained using MAE, M-MAE (ours), U-MAE, and MU-MAE (ours). }
  \label{fig:appdx-pca}
\end{figure*}

\begin{figure*}
  \vspace{-0.5cm}
  \centering
  \includegraphics[width=\textwidth]{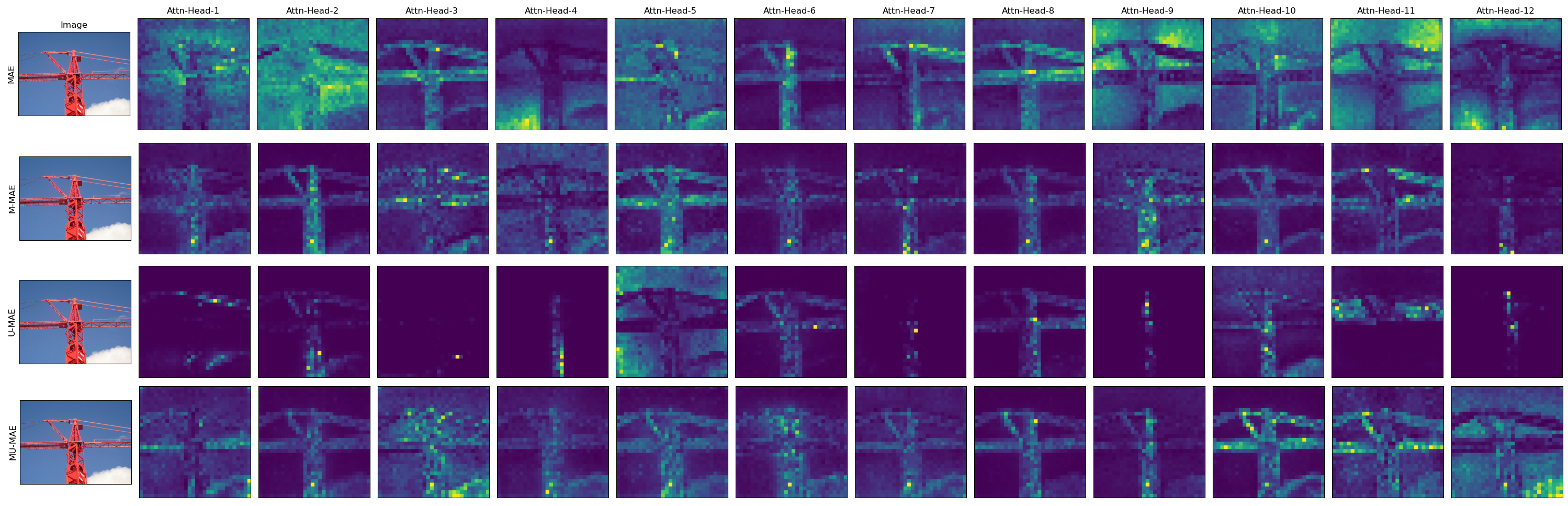}
  \vspace{-0.5cm}
  \caption{Attention maps from the 12 attention heads of the last layer of a ViT-B. The maps are extracted over the four MAE-based methods evaluated: MAE, M-MAE (ours), U-MAE, MU-MAE (ours)}
  \label{fig:appdx-attention}
\end{figure*}

\textbf{Attention maps.} We investigate the impact of the different regularizations on the self-attention maps of the ViT-B architecture's last layer. To this end, we randomly select images from the ImageNet-1K validation set and visualize their corresponding attention maps in \cref{fig:appdx-attention}. Our observations reveal that the baseline MAE model often tends to attend to the background of the image, in line with findings from prior work \cite{lehner2023contrastive}. In contrast, we notice differences when applying \magma{}: it appears to promote a semantic separation of the attention focus, where the model tends to attend primarily to either the background or the central object, but rarely both simultaneously. This suggests that \magma{} guides the model towards learning more specialized and semantically coherent representations, improving its ability to distinguish between foreground and background elements. 

\section{Concluding Remarks}

We propose \magma{} a novel regularization technique that regularizes the representations and enforces consistency across different layers of the transformer-based Masked Autoencoder. We demonstrate the efficacy of the proposed approach through an extensive suite of experimentation resulting in significant performance gain over MAE-based baselines in most scenarios.

\textbf{Computational complexity}. M-MAE offers a 1.5\% and MU-MAE a 2\% drop in throughput (100MB peak GPU-memory) compared to their respective MAE and U-MAE baselines, thus adding a barely noticeable additional computation cost to the baseline methods while keeping the parameter count the same all across. This demonstrates the efficiency and scalability of \magma{}. More details are included in Section 3 of the supplementary material.

\textbf{Broader impact}. \magma{} can be rather straightforwardly applied to any kind of SSL approach irrespective of the backbone architecture. As discussed in Section~\ref{sec:method} this applies to any layered deep networks, irrespective of an encoder-decoder architecture. This potentially broadens the application of \magma{} to contexts even beyond computer vision. This is an avenue for future work.

\textbf{Limitations.} Going beyond ViT-based architectures to CNNs, we observed that the impact of MAGMA is considerably smaller. We argue that standard operations in modern CNN-based architectures (such as average pooling, weight sharing, etc.) might already serve as a regularizer, minimizing the impact of \magma{}. Note that CNN architectures are outside the scope of this work (as reflected throughout the paper), thus, will investigate this in our future work.

%%%%%%%%% REFERENCES
{\small
\bibliographystyle{ieee_fullname}
\bibliography{egbib}
}

\end{document}

\newcommand\blfootnote[1]{%
  \begingroup
  \renewcommand\thefootnote{}\footnote{#1}%
  \addtocounter{footnote}{-1}%
  \endgroup
}
\newcommand{\magma}[0]{\texttt{MAGMA}}
\newcommand{\lreg}[0]{$\mathcal{L}_{\text{Reg}}$}

% \title{\LaTeX\ Author Guidelines for \confName~Proceedings}
\title{Suplementary Material for \\
\texttt{MAGMA:} Manifold Regularization for MAEs}

\author{Alin Dondera$^{*,1}$, Anuj Singh$^{*,1,2}$, Hadi Jamali-Rad$^{1,2}$\\
$^1$Delft University of Technology (TU Delft), The Netherlands \\
$^2$Shell Global Solutions International B.V., Amsterdam, The Netherlands \\
{\tt\small a.e.dondera@student.tudelft.nl, \{a.r.singh, h.jamalirad\}@tudelft.nl}
% \and
}
% \and

\maketitle

%%%%%%%%% BODY TEXT
\begingroup
\begin{table*}[h!]
\setlength{\tabcolsep}{4pt} % Default value: 6pt
  \caption{Sets of differing parameters for MAE, M-MAE, U-MAE, and MU-MAE across the given datasets}
  \label{tab:mae-hyperparameters}
  \centering
  \begin{tabular}{@{}l|cccccc@{}}
    \hline
    Dataset & Backbone & Patch Size & Epochs & Batch Size & lr & $e_{\textup{st}}$ (Reg. warmup) \\
    \hline
    CIFAR-100 & ViT-Tiny & 4 & 2000 & 256 & $1.5e^{-4}$ & 60\\
    Tiny-ImageNet & ViT-Tiny & 8 & 800 & 512 & $1.0e^{-3}$ & 10 \\
    STL-10  & ViT-Tiny & 12 & 800 & 512 & $3.0e^{-4}$ & 10 \\
    ImageNet-100 & ViT-Base & 16 & 400 & 256 & $1.5e^{-4}$ & 10 \\
    \hline
  \end{tabular}
\end{table*}
\endgroup

\begingroup
\begin{table*}
\setlength{\tabcolsep}{4pt} % Default value: 6pt
  \caption{Sets of differing parameters for SimCLR, M-SimCLR, VICReg, and M-VICReg across the given datasets}
  \label{tab:non-generative-hyperparameters}
  \centering
  \begin{tabular}{@{}l|cccccc@{}}
    \hline
    Dataset & Backbone & Patch Size & Epochs & Batch Size & lr & $e_{\textup{st}}$ (Reg. warmup) \\
    \hline
    CIFAR-100 & ViT-Tiny & 4 & 1000 & 256 & $1.0e^{-3}$ & 10 \\
    Tiny-ImageNet & ViT-Tiny & 8 & 1000 & 256 & $1.0e^{-3}$ & 10 \\
    STL-10  & ViT-Tiny & 12 & 1000 & 256 & $1.0e^{-3}$ & 10 \\
    ImageNet-100 & ViT-Tiny & 16 & 200 & 256 & $1.0e^{-3}$ & 10 \\
    \hline
  \end{tabular}
\end{table*}
\endgroup

% \begingroup
% \begin{table*}[!h]
% \setlength{\tabcolsep}{4pt} % Default value: 6pt
%   \caption{Linear probing accuracy of MAE and M-MAE models pretrained on ImageNet-100 across various datasets. Adding \magma{} significantly improves results when evaluated on unseen datasets.}
%   \label{tab:appdx-transfer-learning}
%   \centering
%   \begin{tabular}{@{}l|cccc@{}}
%     \hline
%     Method & CIFAR-100 & STL-10 & Tiny-ImageNet & ImageNet-100\\
%     \hline
%     MAE  & 31.5 & 67.8 & 27.8 & 58.0 \\
%     M-MAE (ours) & \textbf{51.6} & \textbf{84.8} & \textbf{43.1} & \textbf{69.0} \\
%   \hline
%   \end{tabular}
% \end{table*}
% \endgroup

% \begingroup
% \begin{table*}[!h]
% \setlength{\tabcolsep}{4pt} % Default value: 6pt
%   \caption{Finetuning accuracy of MAE and M-MAE models pretrained on ImageNet-100. No significant differences can be seen.}
%   \label{tab:appdx-finetune}
%   \centering
%   \begin{tabular}{@{}l|cccc@{}}
%     \hline
%     Method & CIFAR-100 & STL-10 & Tiny-ImageNet & ImageNet-100 \\
%     \hline
%     MAE  & \textbf{76.9} & 82.2 & \textbf{63.1} & 79.8 \\
%     M-MAE (ours) & 75.6 & \textbf{85.5} & \textbf{63.1} & \textbf{80.6} \\
%   \hline
%   \end{tabular}
% \end{table*}
% \endgroup

\begingroup
\begin{table*}[!h]
\setlength{\tabcolsep}{4pt}
  \caption{Parameter, Throughput in Images/seconds (Img/sec) and GPU Memory for MAE, U-MAE w and w/o regularization.}
  \vspace{-0.3cm}
  \label{tab:profiling}
  \centering
  {\begin{tabular}{@{}cccc@{}}
    \toprule
    Method & Parameters & Img/sec & GPU Memory\\
    \midrule
    MAE  & 122M & 489 & 28.0 GB \\
    M-MAE (ours) & 122M & 481 & 28.1 GB  \\
    \midrule
    U-MAE & 122M & 486 & 28.0 GB \\
    MU-MAE (ours) & 122M & 476 & 28.1 GB \\
  \bottomrule
  \end{tabular}}
\end{table*}
\endgroup

\section{Experimental setup}
\label{sec:appdx-experimental-setup}
\blfootnote{* equal contribution}

\noindent\textbf{Environment details}. \magma{} builds upon the solo-learn \cite{JMLR:v23:21-1155} library of self-supervised methods for unsupervised visual representation learning. All methods are implemented using PyTorch 1.13 and PyTorch Lightning 1.7.7. The following GPUs are used, depending on availability: NVIDIA GeForce RTX 2080 Ti, NVIDIA Tesla V100, and NVIDIA A40.

\noindent\textbf{Datasets.} We conduct our experiments on the following four benchmark datasets:
\begin{itemize}
    \item[--] \textbf{CIFAR-100} \cite{krizhevsky2009learning} consists of 60,000 color images (32x32 pixels) divided into 100 classes, with 500 training images and 100 test images per class. This large number of classes with relatively few images per class pushes models to learn nuanced, discriminative representations for robust classification.
    \item[--] \textbf{STL-10} \cite{coates2011analysis} Contains 5,000 labeled training images, 8,000 test images, and 100,000 unlabeled images (96x96 pixels) across 10 classes. This setting of abundant unlabeled data allows the exploration of self-supervised representation learning techniques, offering a valuable testbed for scenarios where labeled data is scarce.
    \item[--] \textbf{Tiny-ImageNet} \cite{le2015tiny} is a downsized version of ImageNet with 200 classes, featuring 100,000 training images, 10,000 validation images, and 10,000 test images (64x64 pixels). This dataset bridges the gap between smaller benchmarks and full ImageNet, allowing experimentation with larger-scale image recognition tasks while maintaining computational feasibility.
    \item[--] \textbf{ImageNet-100} \cite{tian2020contrastive} is a curated subset of the full ImageNet with approximately 130,000 images (variable resolutions) across 100 classes. It provides a standard train/test split, offering a manageable platform to test the scalability and efficiency of models before moving to the full complexity of ImageNet.
\end{itemize}

This collection of datasets provides a larger range of image classification challenges by varying scales, class complexities, and train/test splits. This suite enables a robust evaluation of the effectiveness of representation learning methods and their generalization across diverse scenarios.

\noindent\textbf{Pretraining hyperparameters}. We split the parameters into three categories: (i) common parameters across all methods and datasets, (ii) parameters used for the MAE-based methods (MAE \cite{he2022masked}, M-MAE, U-MAE \cite{zhang2022mask}, and MU-MAE), (iii) parameters used for SimCLR \cite{tian2020contrastive}, M-SimCLR, VICReg \cite{bardes2021vicreg}, and M-VICReg. The complete configuration files for all combinations of datasets and methods can also be found in the attached code archive.

\noindent\textbf{(i) Common parameters.} All methods use AdamW as an optimizer, with an initial warmup phase of 10 epochs, and an initial learning rate of $3e-5$ decaying to 0 via cosine annealing. Normalization is applied using the specific mean and standard deviation computed across each given dataset.

\noindent\textbf{(ii) MAE-based methods.} Mask ratio for all parameters is 0.75, following \cite{he2022masked}. For U-MAE and MU-MAE, the uniformity weight is set to 0.01, following \cite{zhang2022mask}. The weight for the \magma{} loss is set to 1. For augmentations, we use a random resized crop (scale ranging between $0.08$ and $1$), followed by a random horizontal flip with a probability of $0.5$. The crop is resized to $32\times32$ for CIFAR-100, $64\times64$ for Tiny-ImageNet, $96\times96$ for STL-10, and $224\times224$ for ImageNet-100. All other parameters unrelated to the regularization terms are shared between all methods, and only depend on the dataset. These can be seen in \Cref{tab:mae-hyperparameters}.

\noindent\textbf{(iii) Non-generative SSL methods.} For SimCLR and M-SimCLR we use a temperature of $0.2$. For VICReg and M-VICReg, we use the best weights from \cite{bardes2021vicreg} for the similarity, variance, and covariance loss terms ($25$, $25$, and $1$). The hidden dimensionality of the projector is equal to $2048$ for all. For augmentations, each method follows the parameters described in the original paper. The rest of the relevant parameters can be found in \Cref{tab:non-generative-hyperparameters}.

% \section{Additional evaluations}
% \label{sec:appdx-evaluations}

% We further demonstrate the effectiveness of \magma{} when evaluated on unseen datasets in \Cref{tab:appdx-transfer-learning}, offering substantial improvements over MAE.

% \magma{} is designed to enhance self-supervised representation learning at the pretraining phase. To demonstrate the impact, we keep our supervised fine-tuning strategy as simple as linear probing. Full fine-tuning (especially in low-data regimes) can lead to overfitting to the target dataset, completely defeating the purpose and ruling out the impact of the regularization. That is what we also observe in the new \Cref{tab:appdx-finetune}, with results between the baseline method and ours being close to identical. In \cite{zhang2022mask}, the authors also notice the same effect of fine-tuning on their regularization method.

\section{Computational Complexity}
\label{sec:comp-eff}
We compute the total number of parameters, time efficiency measured by throughput (images per second) and memory efficiency by peak GPU-memory consumption of MAE, UMAE with and without our regularization, and compare them in Table \ref{tab:profiling}. Given the additional cost of computing the sample-wise similarity matrix / laplacian across an entire batch, M-MAE offers a 1.5\% and MU-MAE a 2\% drop in throughput as compared to their respective MAE and U-MAE baselines. This translates to a minor increase of 100MB GPU-memory consumption during training, thus adding an insignificant extra computation cost to the baseline methods. The parameter count of all methods remain the same since our regularization method operates on the same architecture as the MAE and U-MAE baselines.

\section{Additional visualizations}
\label{sec:appdx-visualizations}

\begin{figure*}[h]
  \centering
  \includegraphics[width=\textwidth]{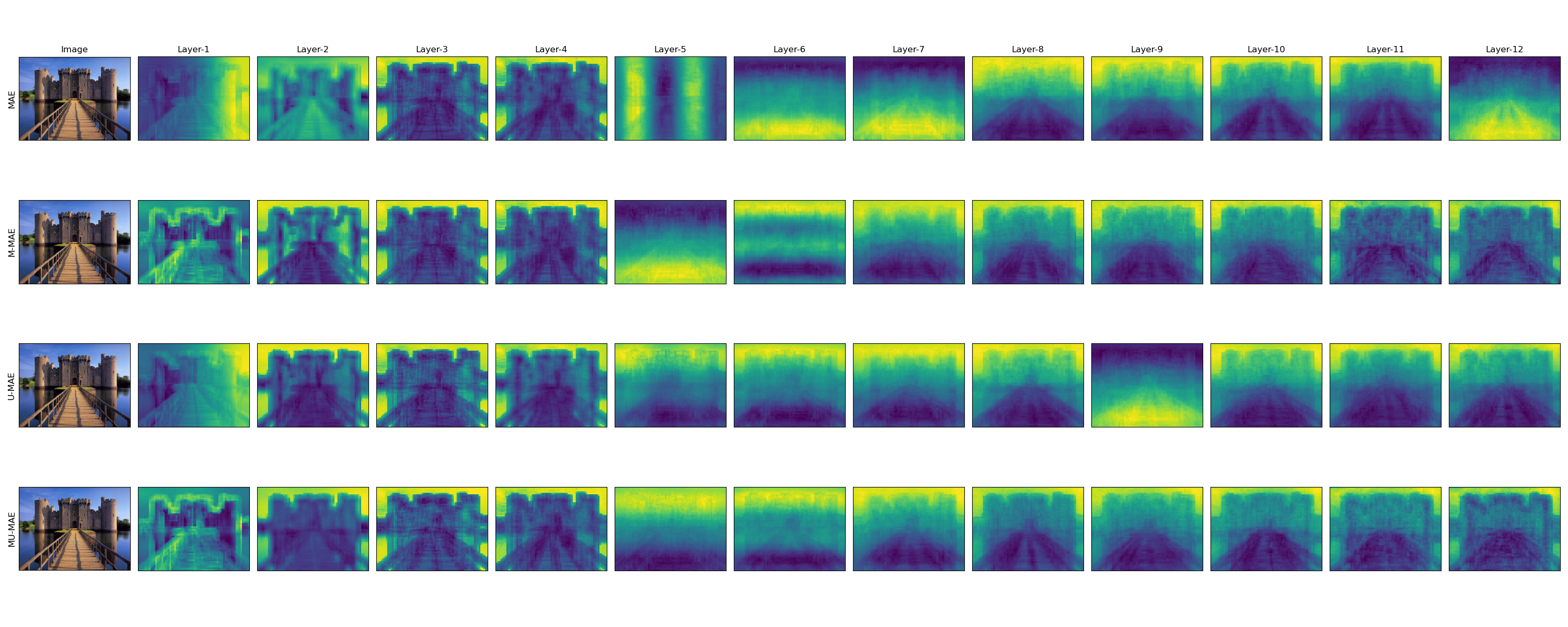}
  \includegraphics[width=\textwidth]{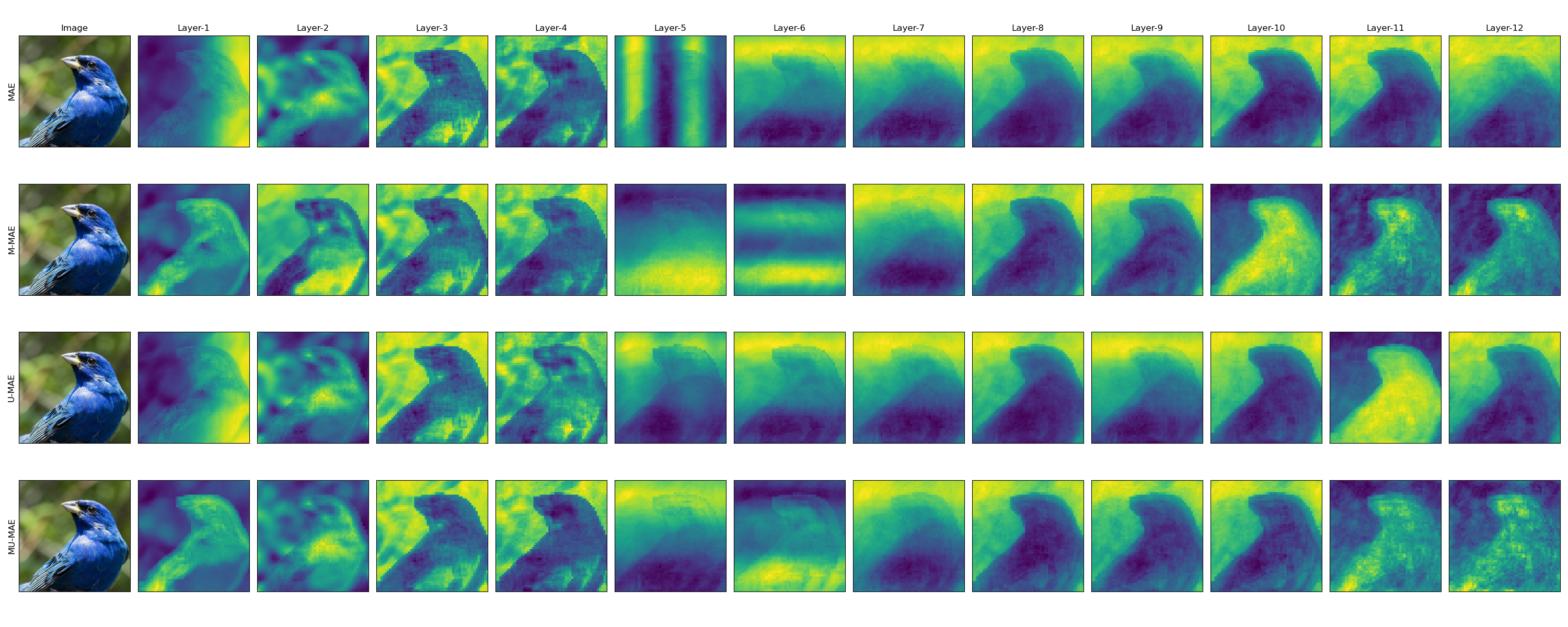}
  \includegraphics[width=\textwidth]{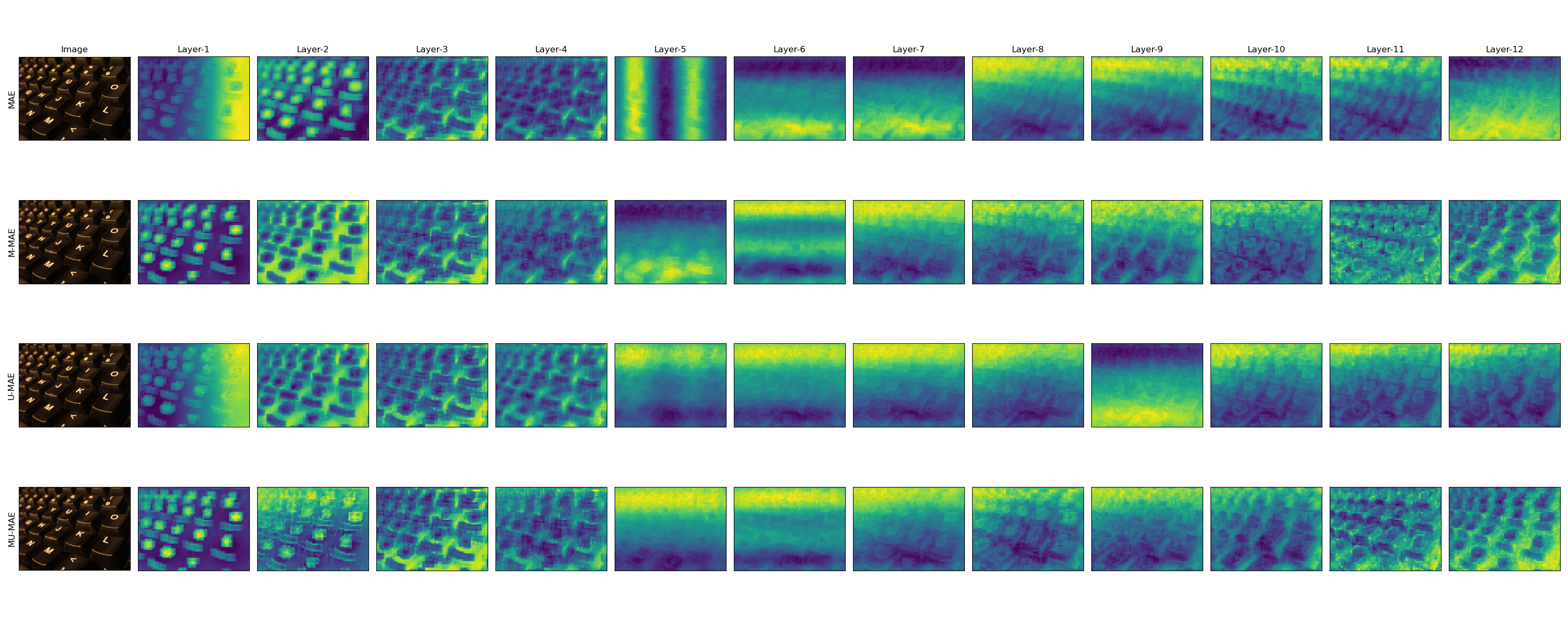}
  \caption{Additional visualization of PCA's leading component for features extracted from different layers of a ViT-B pretrained using MAE, M-MAE (ours), U-MAE, and MU-MAE (ours). }
  \label{fig:appdx-pca}
\end{figure*}

\begin{figure*}[h]
  \centering
  \includegraphics[width=\textwidth]{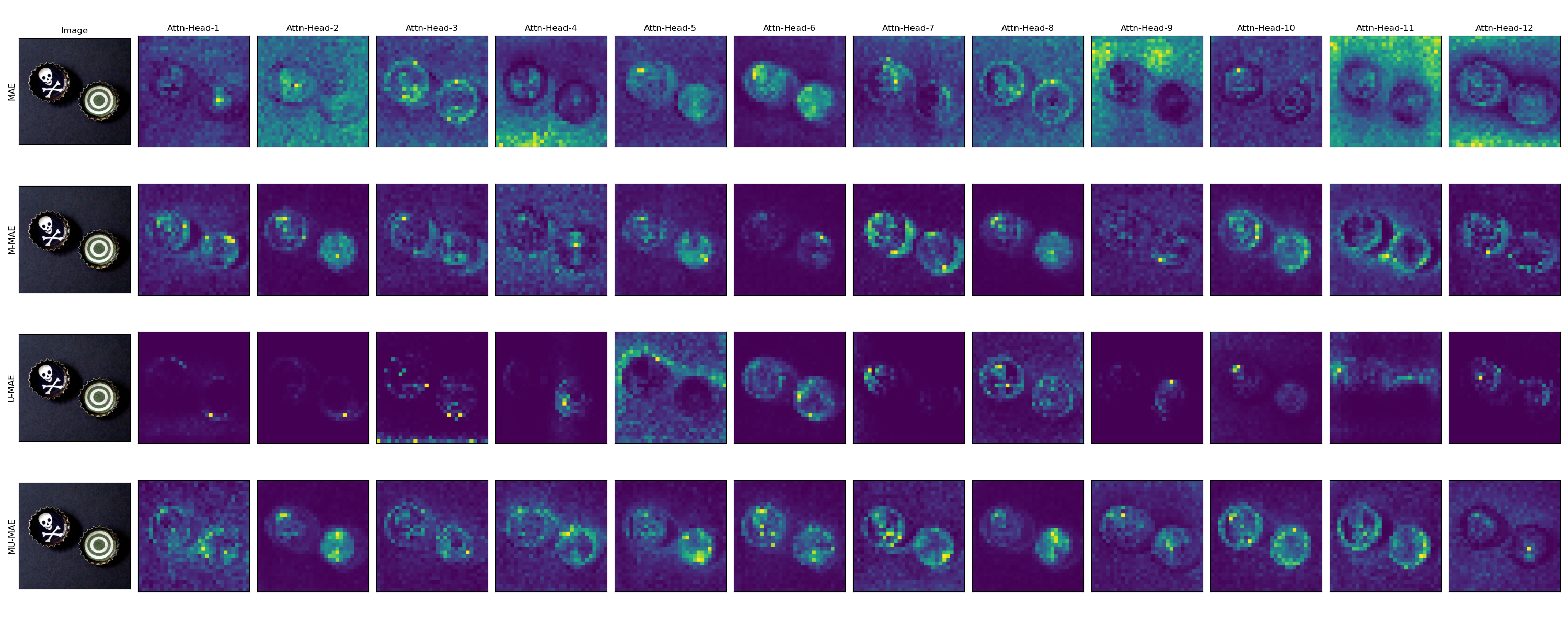}
  \includegraphics[width=\textwidth]{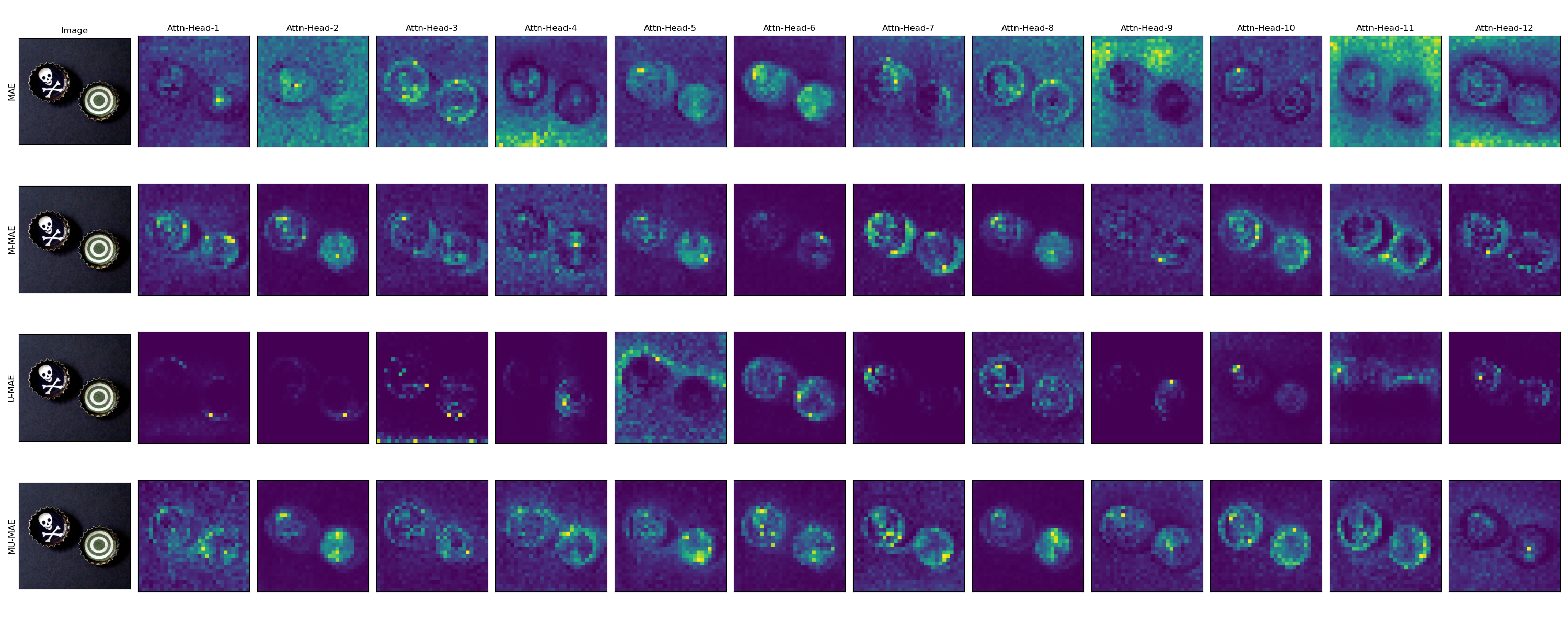}
  \includegraphics[width=\textwidth]{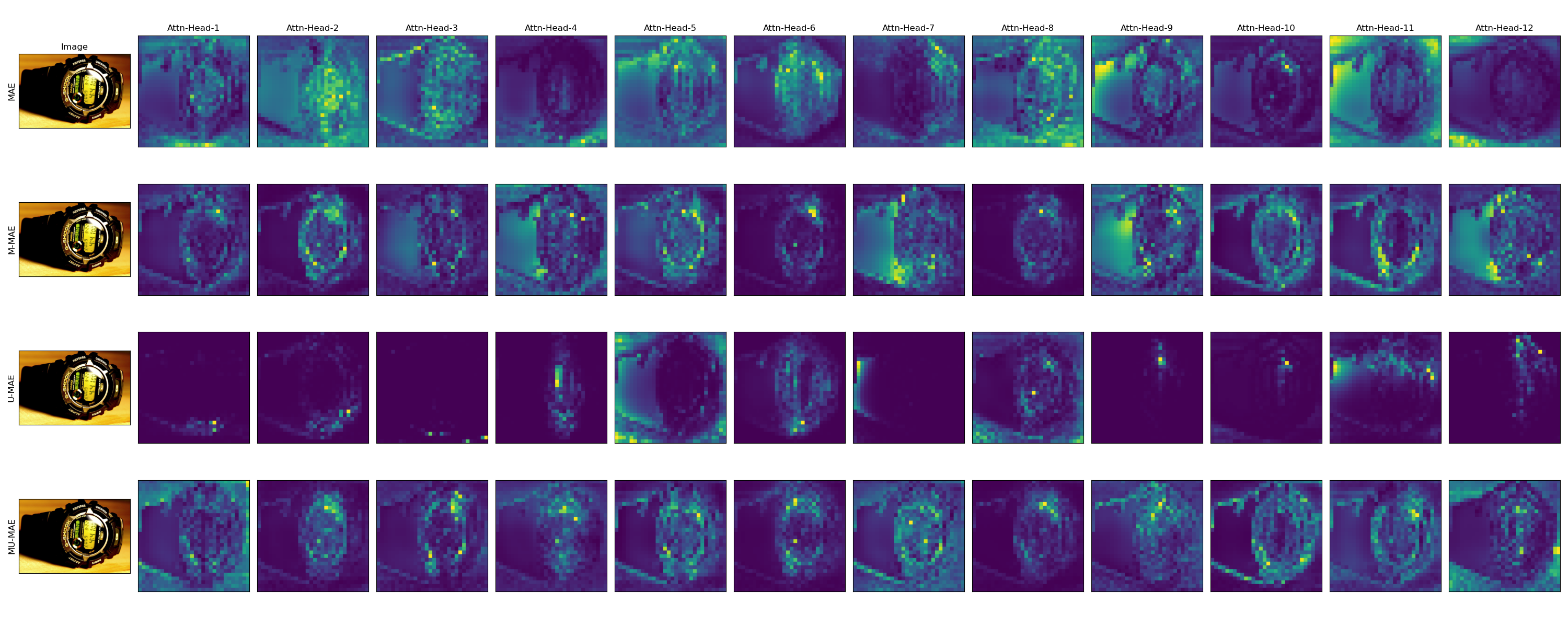}
  \caption{Additional attention maps from the 12 attention heads of the last layer of a ViT-B. The attention maps come from three different images, and for each image, we extract them over the four MAE-based methods evaluated: MAE, M-MAE (ours), U-MAE, MU-MAE (ours)}
  \label{fig:appdx-attention}
\end{figure*}

We provide additional visualizations of the PCA and attention map results in \Cref{fig:appdx-pca} and \Cref{fig:appdx-attention}, as presented in Section 5.4, using a broader selection of images sampled from the ImageNet validation set. These additional visualizations further confirm the previously observed patterns and trends.

{\small
\bibliographystyle{ieee_fullname}
\bibliography{supplementary}
}